\documentclass[letterpaper]{article}
\usepackage{times}
\usepackage[stable]{footmisc}
\usepackage{graphicx}
\usepackage{breakcites}
\usepackage{algorithm,algorithmic,a4wide,amssymb,multicol,enumitem,url}
\usepackage{amsmath}
\usepackage[breaklinks]{hyperref}
 \usepackage[top=3.7cm, bottom=3.7cm, left=3.7cm, right=3.7cm]{geometry} 
\makeatletter
\renewcommand\hyper@natlinkbreak[2]{#1}
\makeatother
\newcommand{\mytilde}{\raise.17ex\hbox{$\scriptstyle\mathtt{\sim}$}}
\begin{document}

\title{One Big Net For Everything \\
{\small Technical Report}
}

\date{24 February 2018 \\ 
{\small Earlier drafts:  23 Aug, 30 Aug, 4 Sep, 31 Oct, 25 Nov, 14 Dec 2017}}

\author{J\"{u}rgen Schmidhuber~\\
The Swiss AI Lab, IDSIA  \\
Istituto Dalle Molle di Studi sull'Intelligenza  Artificiale \\
Universit\`{a} della Svizzera italiana \\
Scuola universitaria professionale della Svizzera italiana \\
Galleria 2, 6928 Manno-Lugano, Switzerland \\ 
%NNAISENSE 
}
\maketitle

\begin{abstract}

I apply recent work on ``learning to think" (2015) ~\cite{learningtothink2015} 
and on {\sc PowerPlay} (2011)~\cite{powerplay2011and13}
to the incremental training of an increasingly general problem solver,
continually 
learning to solve new tasks without forgetting previous skills. 
The problem solver is a single recurrent neural network
(or similar general purpose computer) called ONE.
ONE may sometimes grow or shrink, e.g., 
by adding or pruning neurons and connections,
as proposed in 1965~\cite{ivakhnenko1965,ivakhnenko1971}.
ONE is unusual in the sense that  it is trained
in various ways, e.g., by black box optimization /
reinforcement learning / artificial evolution as well as 
supervised / unsupervised learning.
For example, ONE may learn through neuroevolution to
control  a robot through 
environment-changing actions, and learn through unsupervised gradient descent to
predict future inputs and 
vector-valued reward signals~\cite{Schmidhuber:90diffgenau,Schmidhuber:90sandiego,Schmidhuber:91nips} as suggested in 1990.  
User-given tasks can be defined through extra goal-defining input patterns, also
proposed in 1990~\cite{SchmidhuberHuber:91,Schmidhuber:90compositional,Schmidhuber:91icannsubgoals,SchmidhuberWahnsiedler:92sab}.
Suppose ONE has already learned many skills. 
Now a copy of ONE can be re-trained to learn a new skill,
e.g., through slow trial and error-based neuroevolution without a teacher.
Here it may profit from re-using previously learned subroutines,
but it may also forget previous skills. 
Then ONE 
is  retrained in  {\sc PowerPlay} style (2011)~\cite{powerplay2011and13}
on stored input/output traces of (a) ONE's copy executing the new skill and (b)
previous instances of ONE whose skills are still considered worth memorizing.
Simultaneously, ONE is retrained on old traces (even those of unsuccessful trials)  to become a better predictor.
This is done 
through well-known, feasible,  gradient-based methods,
without additional expensive interaction with the enviroment.
More and more  control and prediction skills are  thus collapsed into ONE, 
like in the chunker-automatizer system of the neural history compressor (1991)~\cite{chunker91and92}.
This forces ONE  to relate partially analogous skills 
(with shared algorithmic information) to each other, 
creating common subroutines in form of shared subnetworks of ONE,
to greatly speed up subsequent learning of 
additional, novel but  algorithmically related skills. 
 
\end{abstract}

%\newpage
%\vspace*{-7\baselineskip}
%\vspace*{4\baselineskip}
%\tableofcontents

\newpage

\section{Introduction}
\label{intro}

I will first quickly summarize a few relevant concepts 
discussed in much more detail in previous reports
\cite{888,learningtothink2015}. The reader might  profit from being familiar 
with some of our earlier work on 
algorithmic transfer learning~\cite{Schmidhuber:04oops,powerplay2011and13,learningtothink2015} and 
recurrent neural networks (RNNs) 
for control and  planning~\cite{Schmidhuber:90diffgenau,Schmidhuber:90sandiego,Schmidhuber:91nips,SchmidhuberHuber:91} and 
hierarchical chunking~\cite{chunker91and92}.

To become a general problem solver that is able to run arbitrary
problem-solving programs, the controller of a robot or an artificial
agent must be a general-purpose
computer~\cite{Goedel:31,Church:36,Turing:36,Post:36}.  Artificial RNNs fit this bill.  A typical RNN
consists of many simple, connected processors called neurons, each
producing a sequence of real-valued activations.  Input neurons get
activated through sensors perceiving the environment, other neurons
get activated through weighted connections or wires from previously
active neurons, and some neurons may affect the environment by
triggering actions.  {\em Learning} or {\em credit assignment} is
about finding real-valued weights that make the RNN exhibit {\em
  desired} behavior, such as driving a car.  
   The weight matrix of an RNN is
its program.  

Many RNN-like models can be used to build general computers, e.g.,
RNNs controlling pushdown automata~\cite{Das:92,mozer1993connectionist} 
or other types of differentiable memory~\cite{graves2016}
including differentiable fast
weights~\cite{Schmidhuber:92ncfastweights,Schmidhuber:93ratioicann}, as well as closely related
RNN-based meta-learners~\cite{Schmidhuber:93selfreficann,Hochreiter:01meta,scholarpedia2010}.
Using sloppy but convenient terminology, we refer to all of them as
RNNs~\cite{learningtothink2015}.  In practical applications, most RNNs are {\em Long Short-Term Memory} (LSTM)
networks~\cite{lstm97and95,Gers:2000nc,Graves:09tpami,888},
now used billions of times per day for automatic translation~\cite{wu2016google,facebook2017}, speech recognition~\cite{googlevoice2015},  and many other tasks~\cite{888}. 
If there are large 2-dimensional inputs such as
video images, the LSTM may have a front-end~\cite{vinyals2014caption} in form of  a convolutional neural net  
(CNN)~\cite{Fukushima:1979neocognitron,LeCun:89,weng1992,Behnke:LNCS,ranzato-cvpr-07,scherer:2010,ciresan2012cvpr,888}
implemented on fast GPUs~\cite{ciresan2012cvpr,888}.
Such a
CNN-LSTM combination is still an RNN.

Without a teacher, reward-maximizing programs of an RNN
must be learned through repeated trial and error,
e.g., through artificial evolution~\cite{miller:icga89,yao:review93,Sims:1994:EVC,moriarty:phd,gomez:phd,Gomez:03,wierstraCEC08,glasmachers:2010b,Sun2009a,sun:gecco13}
\cite[Sec.~6.6]{888}, or reinforcement learning~\cite{Kaelbling:96,Sutton:98,wiering2012,888}
through policy gradients~\cite{Williams:86,Sutton:99,baxter2001,aberdeenthesis,ghavamzadehICML03,stoneICRA04,wierstraCEC08,rueckstiess2008b,sehnke2009parameter,gruettner2010multi,wierstra2010,peters2010}
\cite[Sec.~6]{888}.
The search space can often be reduced dramatically by evolving
{\em compact encodings} of RNNs, e.g.,~\cite{Schmidhuber:97nn+,stanley09,koutnik:gecco13,steenkiste2016wavelet}\cite[Sec.~6.7]{888}. 
Nevertheless, this is often much harder than imitating teachers
through gradient-based supervised learning
\cite{Werbos:88gasmarket,WilliamsZipser:92,RobinsonFallside:87tr}\cite{888}
for  LSTM~\cite{lstm97and95,Gers:2000nc,Graves:09tpami}. 

However, reinforcement learning RNN controllers can profit from gradient-based RNNs used as predictive 
world models~\cite{Schmidhuber:90diffgenau,Schmidhuber:90sandiego,Schmidhuber:91nips,learningtothink2015}.
See previous papers for many additional references on this~\cite{888,learningtothink2015}.
In what follows, I will elaborate on  such previous work.

\section{One Big RNN For Everything: Basic Ideas and Related Work}
\label{ONE}

I will focus on
the  incremental training of an increasingly general problem solver
interacting with an environment, continually~\cite{Ring:94}
learning to solve new tasks (possibly without supervisor) and without forgetting any previous, still valuable skills. 
The problem solver is a single RNN called ONE.

Unlike previous RNNs, 
 ONE or copies thereof or parts thereof are trained
in various ways, in particular, by {\bf (1)} black box optimization /
reinforcement learning / artificial evolution without a teacher, 
or {\bf (2)} gradient descent-based supervised  or 
unsupervised learning (Sec. \ref{intro}).
{\bf (1)} is usually much harder than {\bf (2)}.
Here I combine {\bf (1)} and {\bf (2)} in a way that leaves much if not most of the work to {\bf (2)},
building on several ideas from previous work:

\begin{enumerate}[leftmargin=*]
\item {\bf Extra goal-defining input patterns to encode user-given tasks.} 
A reinforcement learning neural controller of 1990 learned to control a fovea through 
sequences of saccades to find particular objects in visual scenes,
thus learning sequential attention~\cite{SchmidhuberHuber:91}. User-defined goals were 
provided to the system by special ``goal input vectors" that remained constant 
\cite[Sec.~3.2]{SchmidhuberHuber:91}
while
the system shaped its incoming stream of standard 
visual inputs through its fovea-shifting actions.
Also in 1990, gradient-based recurrent subgoal generators 
~\cite{Schmidhuber:90compositional,Schmidhuber:91icannsubgoals,SchmidhuberWahnsiedler:92sab}
used special start and goal-defining input vectors, also for an evaluator network predicting the costs and rewards
associated with moving from starts to goals. 
The later  {\sc PowerPlay} system (2011)~\cite{powerplay2011and13} also used such task-defining
special inputs, actually selecting on its own new goals and tasks, to become
a more and more general problem solver in an active but unsupervised fashion. 
In the present paper, variants of 
ONE will also adopt this concept of extra goal-defining inputs to distinguish between numerous different tasks.

\item {\bf Incremental black box optimization of reward-maximizing RNN controllers}. If ONE already knows
how to solve several tasks, then a copy of ONE may profit from this prior knowledge, 
learning a new task through additional weight changes more quickly
than learning the task from scratch, e.g.,~\cite{gomez:ab97,stoneML05,ghavamzadehICML03},
ideally through  optimal {\em algorithmic transfer learning}, like in the
at least  
 {\em asymptotically Optimal Ordered Problem Solver}~\cite{Schmidhuber:04oops},
where new solution candidates in form of programs may exploit older ones in arbitrary computable fashion. 

\item {\bf Unsupervised prediction and compression of all data of all trials.}  
An RNN-based ``world model" M 
of 1990~\cite{Schmidhuber:90diffgenau,Schmidhuber:90sandiego}  learned to predict (and thus compress~\cite{chunker91and92}) 
future inputs including 
vector-valued reward signals~\cite{Schmidhuber:90sandiego}
from the environment of an agent controlled by another RNN called C through 
environment-changing actions. This was also done in more recent, 
more sophisticated CM systems~\cite{learningtothink2015}. 
Here we collapse both M and C into ONE, 
very much like in Sec. 5.3 of the previous paper~\cite{learningtothink2015},
where C and M were bi-directionally connected such that they effectively became one big net 
that ``learns to think"~\cite{learningtothink2015}.
In the present paper, however, we do not make any explicit difference any more between C and M. 

\item {\bf Compressing all behaviors so far into ONE.}  
The chunker-automatizer system of the neural history compressor of 
1991~\cite{chunker91and92,schmidhuber1993} 
used gradient descent to compress the learned behavior of a so-called  {\em ``conscious"} chunker RNN
into a separate {\em ``subconscious"} automatizer RNN, 
which not only learned to imitate the chunker network,
but also was continually retrained on its own previous tasks,
namely, (1) to predict teacher-given targets through supervised learning, 
and (2) to compress through unsupervised learning all sequences of observations by predicting them (what is predictable does not have to be stored extra). 
It was shown that this type of unsupervised pretraining for deep learning networks can greatly facilitate 
the learning of additional user-defined tasks~\cite{chunker91and92,schmidhuber1993}.

Here we apply the basic  idea to the  incremental skill training of ONE. 
Both the predictive skills acquired by gradient descent and the task-specific control skills acquired by black box optimization
can be collapsed into one single network (namely, ONE itself) through pure gradient descent,
by retraining ONE on all input-output traces of all previously learned
behaviors that are still deemed useful~\cite{powerplay2011and13}. 
Towards this end, we simply retrain ONE to reproduce control behaviors of successful past versions of ONE,
but without really executing the behaviors in the environment (usually the expensive part).
Simultaneously, all input-output traces ever observed (including those of failed trials) can be used
 to train ONE to become a better predictor of future inputs,
 given previous inputs and actions.
Of course, this requires to store input-output traces of all trials~\cite{Schmidhuber:06cs,Schmidhuber:09sice,learningtothink2015}.

\end{enumerate}

That is,
once a new skill has been learned by a copy of ONE
(or even by another machine learning device),
e.g., through slow trial and error-based evolution or reinforcement learning,
ONE is simply retrained in  {\sc PowerPlay} style~\cite{powerplay2011and13}
through well-known, feasible,  
{\em gradient-based} methods on stored input/output traces~\cite[Sec.~3.1.2]{powerplay2011and13}
of all previously learned control and prediction skills still considered worth memorizing, 
similar to the chunker-automatizer system of the neural history compressor of 
1991~\cite{chunker91and92}. 
In particular, standard 
gradient descent through 
backpropagation  in discrete graphs of
nodes with differentiable activation 
functions~\cite{Linnainmaa:1970,Werbos:81sensitivity}\cite[Sec.~5.5]{888}
can be used to squeeze many expensively evolved skills into the limited computational 
resources of ONE.
Compare recent work on incremental skill learning~\cite{progressive2018}.
Well-known regularizers \cite[Sec.~5.6.3]{888} can be used to further compress ONE, 
possibly shrinking it by pruning neurons and connections,
as proposed already in 1965 for deep learning multilayer perceptrons~\cite{ivakhnenko1965,ivakhnenko1971,learningtothink2015}.
This forces ONE even more  to relate partially analogous skills 
(with shared algorithmic information~\cite{Solomonoff:64,Kolmogorov:65,Chaitin:66,Levin:73a,Solomonoff:78,LiVitanyi:97,Schmidhuber:04oops}) to each other, 
creating common sub-programs in form of shared subnetworks of ONE.
This may greatly speed up subsequent learning of 
novel but  algorithmically related skills,
through reuse of such subroutines created as by-products of data compression,
where the data are actually programs encoded in ONE's previous weight matrices.

So ONE continually collapses more and more skills and predictive knowledge into itself, 
compactly encoding shared algorithmic information in re-usable form,
to learn new problem-solving programs more quickly.

%\newpage

\section{More Formally: ONE and its Self-Acquired Data}
\label{formally}

The notation below is similar but not identical to the one in previous work
on an RNN-based CM system called the RNNAI~\cite{learningtothink2015}. 

Let $m,n,o,p,q,s$ denote positive integer constants, and
$i,k,h,t,\tau$ positive integer variables assuming ranges implicit
in the given contexts.  The $i$-th component of any real-valued vector,
$v$, is denoted by $v_i$.  
For convenience, let us assume that ONE's life span can be partitioned 
into trials $T_1, T_2, \ldots$  In each trial, ONE attempts to solve a particular task,
trying to manipulate some unknown environment through a sequence of
actions to achieve some goal.
Let us consider one particular trial $T$ and its discrete sequence of
time steps, $t = 1,2,\ldots, t_{T}$.

At the beginning of a given time step, $t$, ONE receives a ``normal'' sensory
input vector, $in(t) \in \mathbb{R}^m$, and a reward input vector, $r(t)
\in \mathbb{R}^n$.  For example, parts of $in(t)$ may represent the pixel
intensities of an incoming video frame, while components of $r(t)$ may
reflect external positive rewards, or negative values produced by pain
sensors whenever they measure excessive temperature or pressure or low battery load (hunger). 
Inputs $in(t)$ may also encode user-given goals or tasks, 
e.g., through commands spoken by a user. 
Often, however, it is convenient to use an extra input vector
$goal(t) \in \mathbb{R}^p$ to uniquely encode user-given goals,
as we have done since 1990, e.g.,~\cite{SchmidhuberHuber:91,powerplay2011and13}.
Let
$sense(t) \in \mathbb{R}^{m+p+n}$ denote the concatenation of the
vectors $in(t)$, $goal(t)$  and $r(t)$.  The total reward at time $t$ is $R(t)=
\sum_{i=1}^{n}r_i(t)$.  The total cumulative reward up to time $t$ is
$CR(t)= \sum_{\tau=1}^{t}R(\tau)$.  During time step $t$, ONE
computes during several micro steps (e.g.,~\cite[Sec.~3.1]{learningtothink2015}) an output action vector, $out(t) \in \mathbb{R}^o$, which may
influence the environment and thus future $sense(\tau)$ for $\tau >t$.

\subsection{Training a Copy of ONE on New Control Tasks Without a Teacher}
\label{control}

One of ONE's goals is to maximize $CR(t_{T})$.
Towards this end, copies of successive instances of 
ONE  are trained in a series of trials through
a black box optimization method
in  Step 3 of Algorithm~\ref{ONEalg},
e.g., through
 incremental neuroevolution~\cite{gomez:ab97},
hierarchical  neuroevolution~\cite{stoneML05,vanhoorn:09cig},
hierarchical policy gradient
algorithms~\cite{ghavamzadehICML03},
or  asymptotically optimal ways of {\em algorithmic transfer learning}~\cite{Schmidhuber:04oops}.
Given a new task and a ONE trained on several previous tasks, such
hierarchical/incremental methods may create a copy of the current ONE,  freeze its current weights, 
then enlarge the copy of ONE by adding a few new units and connections~\cite{ivakhnenko1971} which are
trained until the new task is satisfactorily solved.  This process can reduce the size of the search
space for the new task,  while giving the new weights the opportunity to
learn to somehow use certain frozen parts of ONE's copy as subroutines.
(Of course, it is also possible to simply retrain {\em all} weights of the entire copy to solve the new task.)
Compare a recent study of incremental skill learning with feedforward networks~\cite{progressive2018}.

In non-deterministic or noisy environments, 
by definition 
the task is considered solved once the latest version of 
the RNN has performed satisfactorily on a statistically significant numer of trials
according to a user-given criterion, which also implies that
the input-output traces of these trials (Sec.~\ref{collapse}) are sufficient  
to retrain ONE in Step 4 of Algorithm~\ref{ONEalg}
without further interaction with the environment.

\subsection{Unsupervised ONE Learning to Predict/Compress Observations}
\label{compress}

ONE may further profit from 
unsupervised learning that compresses the observed data~\cite{chunker91and92}
into a compact representation that may make subsequent  learning of externally posed tasks easier~\cite{chunker91and92,learningtothink2015}.
Hence, another goal of ONE can be to compress ONE's entire growing
interaction history of all failed and successful trials~\cite{Schmidhuber:06cs,Schmidhuber:10ieeetamd}, e.g., through neural predictive coding~\cite{chunker91and92,SchmidhuberHeil:96}.
For this purpose, ONE has  $m+n$ special output units to produce for $t<t_{T}$
a prediction $pred(t)
\in \mathbb{R}^{m+n}$ of $sense(t+1)$
\cite{Schmidhuber:90diffgenau,Schmidhuber:90sandiego,Schmidhuber:90sab,Schmidhuber:90cmss,Schmidhuber:91nips}
from ONE's previous observations and actions, which are in principle accessible to ONE through (recurrent) connections. 
In one of the simplest cases, this contributes $\| pred(t) - sense(t +1) \|^2$
to the error function to be
minimized by gradient descent in ONE's weights,
in Step 4 of Algorithm~\ref{ONEalg}.
This will train $pred(t)$ to become more like the expected value of of $sense(t+1)$, given the past. 
See previous papers~\cite{SchmidhuberHeil:96,Schmidhuber:06cs,learningtothink2015} for 
ways of translating such neural predictions into compression performance. 
(Similar prediction tasks could also be specified through particular 
prediction task-specific goal inputs $goal(t)$, like with other tasks.)

\subsection{Training ONE to Predict Cumulative Rewards}
\label{cumulative}

We may give ONE yet another set of 
$n$ special output units to produce for $t<t_{T}$ another prediction $PR(t)
\in \mathbb{R}^{n+1}$ of $r(t+1)+r(t+2)+\ldots + r(t_{T})$ and of the 
total remaing reward $CR(t_{T})-CR(t)$~\cite{Schmidhuber:90diffgenau}.
Unlike in the present paper, predictions of 
expected cumulative rewards  are actually {\em essential} in {\em traditional} reinforcement learning~\cite{Kaelbling:96,Sutton:98,wiering2012,888} where they are usually 
limited to  the case of {\em scalar} rewards
(while ONE's rewards may be {\em vector-valued} like in old work of 1990~\cite{Schmidhuber:90diffgenau,Schmidhuber:90sandiego}). 
Of course, in principle, such cumulative knowledge is already 
implicitly present in a ONE that has learned to
predict only next step rewards  $r(t+1)$.
However,
explicit predictions of expected cumulative rewards may represent redundant but useful derived  secondary features that further 
facilitate black box optimization in later incarnations of Step 3  of Algorithm~\ref{ONEalg},
which may discover useful subprograms of the RNN making 
good use of those features.

\subsection{Adding Other Reasonable Objectives to ONE's Goals}
\label{other}

We can add additional objectives to ONE's goals.  For example, 
we may give ONE another set of 
$q$ special output units and train them through unsupervised learning~\cite{Schmidhuber:92ncfactorial}
 to produce for $t \leq t_{T}$ a vector $code(t)
\in \mathbb{R}^{q}$ that represents an ideal factorial code~\cite{Barlow:89review} of the observed history so far,
or that encodes the data in related ways generally considered useful, e.g.,~\cite{herault1984reseau,Jutten:91,Schuster:92,Schmidhuber:99zif,greff2017neural}. 

\subsection{No Fundamental Problem with Bad Predictions of Inputs and Rewards}
\label{bad}

Note that like in work of 2015~\cite{learningtothink2015} but 
unlike in earlier work on {\em learning to plan} of 1990~\cite{Schmidhuber:90diffgenau,Schmidhuber:90sandiego}, 
it is not that important that ONE becomes a good predictor of  inputs
(Sec.~\ref{compress})  including cumulative rewards (Sec.~\ref{cumulative}). 
In fact, in noisy environments, perfect prediction is impossible. 
The learning of solutions of control tasks in  Step 3 of Algorithm~\ref{ONEalg}, however, does not essentially  {\em depend} on good predictions,
although it might profit from internal subroutines of ONE 
(learned in Step 4) that at least occasionally 
yield good predictions of expected future observations
 in form of of $pred(t)$ or $PR(t)$. 

Likewise, control learning may profit from but 
does not existentially  {\em depend} on near-optimal codes 
according to  Sec.~\ref{other}. 

To summarize, ONE's subroutines for making codes and predictions
may or may not help to solve control  problems during Step 3, where 
 it is  ONE's task to figure out when to use or ignore those subroutines.

\subsection{Store Behavioral Traces}
\label{store}

Like in previous work since 2006~\cite{Schmidhuber:06cs,Schmidhuber:09sice,learningtothink2015}, 
to be able to retrain ONE on all observations ever made,
 {\em
  we should store ONE's entire, growing, lifelong sensory-motor interaction
  history  including all inputs and goals and actions and reward signals observed
  during all successful and failed trials~\cite{Schmidhuber:06cs,Schmidhuber:09sice,learningtothink2015},
  including what initially looks like noise but later may turn out to
  be regular}. This is normally not done, but feasible today. 
  Remarkably, as pointed out in 2009, even human brains may have enough storage capacity
to store 100 years of sensory input at a reasonable resolution~\cite{Schmidhuber:09sice}.

On the other hand, in some applications, storage space is limited, and we might want to store (and re-train on)
only some (low-resolution variants) of the previous observations, selected according to certain user-given criteria.
This does not fundamentally change the basic setup - ONE may still profit from subroutines
that encode such limited previous experiences, as long as they convey algorithmic
information about solutions for new tasks to be learned.

\subsection{Incrementally Collapse All Previously Learned Skills into ONE}
\label{collapse}

Let $all(t)$ denote the concatenation of $sense(t)$ and $out(t)$ and $pred(t)$ (and possibly $PR(t)$ and $code(t)$ if any).  
Let $trace(T)$ denote the sequence $(all(1), all(2), \ldots, all(t_{T}))$. 
To combine the objectives of  the previous, very general 
papers~\cite{powerplay2011and13,learningtothink2015},
we can use  
simple, well understood, rather efficient,  {\em gradient-based learning} to compress~\cite{chunker91and92}
all relevant aspects of 
$trace(T_1), trace(T_2), \ldots$  into ONE, and thus compress all 
control~\cite{rusu2016progressive} and prediction~\cite{chunker91and92} skills learned so far by previous instances of ONE (or even by separate machine learning methods),
preventing ONE not only from forgetting previous knowledge,
but also making ONE discover new relations and analogies and
other types of mutual algorithmic information
among subroutines implementing previous skills.  
Typically, given a ONE that already knows many skills, 
traces of a new skill learned by a copy of ONE are added to the relevant traces, 
and compressed into ONE, which is also re-trained on traces of the previous skills.  
See Step 4 of Algorithm~\ref{ONEalg}.

Note that 
 {\sc PowerPlay}  (2011)~\cite{powerplay2011and13,Srivastava2013first} also 
uses environment-independent replay of behavioral traces (or functionally equivalent but more efficient methods)  to avoid forgetting and to compress or speed up
previously found, sub-optimal solutions.  At any given time, an acceptable 
(possibly self-invented)
task is to solve a previously solved task with fewer computational 
resources such as time, space, energy, as long as this does not worsen
performance on other tasks. 
In the present paper, we focus on pure gradient descent for ONE (which may have an LSTM-like architecture)  to implement the  {\sc PowerPlay} principle.

\subsection{Learning Goal Input-Dependence Through Compression}
\label{goals}

After Step 3 of Algorithm~\ref{ONEalg}, a copy of 
ONE may have been modified and may have learned to control an agent in a video game such that it reaches a given goal in a maze,
indicated through a particular goal input, e.g., one that looks a bit like the goal~\cite[Sec.~3.2]{SchmidhuberHuber:91}.
However, the weight changes of ONE's copy may be insufficient to perform this behavior 
{\em exclusively} when the corresponding goal input is on. 
And it may have forgotten previous skills for finding other goals,
given other goal inputs. Nevertheless, the gradient-based~\cite{rusu2016progressive} dreaming phase of Step 4
can correct and fine-tune all those behaviors,
making them goal input-dependent in  a way that would be hard for
typical black box optimizers such as neuroevolution.

\begin{algorithm}[H]
\begin{algorithmic}

\STATE {\bf 1.} 
Access global variables (also accessible to calling procedures such as Algorithm~\ref{simplealg}): 
the present ONE and its weights, positive real-valued variables $c, \lambda$ defining search time budgets,
and a control task description $ A \in \cal T$ 
from a possibly infinite set of possible task descriptions $\cal T$ 
~\cite[Sec.~2]{powerplay2011and13}.

\STATE {\bf 2.} 
Unless goal descriptions are transmitted through normal input units,
e.g., in form of speech,
select  a unique, task-specific~\cite{SchmidhuberHuber:91}
goal input $G(A) \in \mathbb{R}^p$ for ONE;
otherwise $G(A)$ is a vector of $p$ zeros. 

\STATE {\bf 3 (Try to Solve New Task).} 
Make a copy of the present ONE and call it ONE1;
make a copy of the original ONE (before training) and call it ONE0
(notation in both cases like for ONE; Sec. \ref{formally}).
The total search time budget~\cite{Schmidhuber:04oops} of the present Step 3 is $c$ seconds.  
In parallel (or interleaving) fashion,
apply a trial-based black box optimization method (Sec. \ref{control})
to (all or some of the weights of) ONE0 and ONE1, 
spending equal time on both,
until $c$ seconds have been spent without success (then go to Step 4), or until either ONE0 or ONE1 
have learned task $A$ sufficiently well, according to some given termination criterion,
where for both ONE0 and ONE1
for all time steps $t$ of all trials,
$G(A)=goal(t)=const.$ 
In case of first success through ONE0, rename it ONE1. 
If both ONE1 and the environment are deterministic,  such that trials are repeatable exactly,
mark only the final ONE1's $trace(T)$ (Sec.~\ref{collapse}) as {\em relevant}, where $T$ is the final successful trial.   
Otherwise, to gain statistical significance, mark as {\em relevant} the 
traces of sufficiently many (Sec.~\ref{control}) successful trials conducted by the final 
ONE1 on task $A$.

{\em Comment: Previously learned programs and subroutines already 
encoded in the weight matrix of ONE at the beginning of Step 3 
 may help to greatly speed up ONE1's optimization  process - see Sec.~\ref{control}.
ONE0, however, is trying to learn $A$ from scratch, playing the role of a safety belt in case
ONE1 has become ``too biased" through previous learning (following the algorithmic transfer learning approach of the
asymptotically Optimal Ordered Problem Solver~\cite{Schmidhuber:04oops}).}

\STATE {\bf 4 (Dream and Consolidate).} 
Since ONE1 may have forgotten previous skills in Step 3,
and may not even have understood the goal input-dependence of the newly learned behavior for  $A$ (Sec.~\ref{goals}),
spend  $\lambda c$ seconds on:
retrain ONE by {\em standard gradient-based learning} (Sec.~\ref{intro},\ref{collapse})  to reproduce the input history-dependent outputs $out(t)$ in all traces of all previously learned 
{\em relevant}
behaviors that are still deemed useful (including those for the most recent task $A$ learned by ONE1, if any).  
Simultaneously, use all traces (including those of failed trials)
 to retrain ONE  to make better predictions $pred(t)$ (Sec.~\ref{compress})  and  $code(t)$ (Sec.~\ref{other}) if any,
 given previous inputs and actions 
 (but do not provide any target values for action outputs $out(t)$ and corresponding
$PR(t)$ (Sec.~\ref{cumulative}) in replays of formerly {\em relevant} traces of trials of unsuccessful or superseded controllers implemented by earlier incarnations of ONE - see Sec.~\ref{discard}).
Use regularizers to compactify and simplify ONE as much as possible~\cite{888,learningtothink2015}. 

{\em Comment: This process collapses all previous prediction skills and still relevant goal-dependent control skills into ONE,
without requiring new expensive interactions with the environment. We may call this a
consolidation phase or sleep phase~\cite{Schmidhuber:09abials}  
or dream phase or regularity detection phase.}

\end{algorithmic}
\caption{How ONE can learn (without a teacher) one more control skill as well as additional prediction skills, 
using pure gradient-based learning for avoiding to forget previously learned skills and for learning goal input-dependent behavior. See Sec. \ref{formally} for details of steps 3-4.}
\label{ONEalg}
\end{algorithm}

\newpage

The setup is also sufficient for high-dimensional spoken commands 
arriving as input vector sequences at certain standard input units connected to a microphone. 
The non-trivial pattern recognition required to recognize commands such as 
{\em ``go to the north-east corner of the maze"}
will require a substantial subnetwork of ONE and many weights. 
We cannot expect neuroevolution to learn such speech recognition within reasonable time. 
However,  a copy of ONE may rather easily learn by neuroevolution during Step 3 of Algorithm~\ref{ONEalg}
to always go to the north-east corner of the maze, ignoring speech inputs. 
In a later incarnation of Step 3, 
a copy of another instance of ONE may rather easily learn 
to always go to the north-west corner of the maze, again ignoring corresponding spoken commands such as
{\em ``go to the north-west corner of the maze."}
In the consolidation phase of Step 4, ONE then may rather easily learn~\cite{fernandez:icann2007,googlevoice2015} 
the speech command-dependence of these behaviors through gradient-based learning,
without having to interact with the environment again.
Compare the concept of {\em input injection}~\cite{progressive2018}.

\subsection{Discarding Sub-Optimal Previous Behaviors}
\label{discard}

Once ONE has learned to solve some  control task in suboptimal fashion, 
it may later learn to solve it faster, or with fewer computational resources. 
That's why 
Step 4 of Algorithm~\ref{ONEalg} 
does not retrain ONE to generate action outputs $out(t)$
 in replays~\cite{Lin:91} of formerly {\em relevant} traces of trials of superseded controllers implemented by earlier versions of ONE.
However, replays of unsuccessful trials can still be  used to retrain ONE to become a better predictor or world model~\cite{learningtothink2015}, given past observations and actions (Sec.~\ref{compress}).

\subsection{Algorithmic Information Theory (AIT) Argument}
\label{ait}

As discussed in earlier work~\cite{learningtothink2015},
according to the Theory of Algorithmic Information (AIT) or Kolmogorov Complexity~\cite{Solomonoff:64,Kolmogorov:65,Chaitin:66,Levin:73a,Solomonoff:78,LiVitanyi:97}, 
given some universal computer, $U$, whose programs are
encoded as bit strings, the mutual information between two programs
$p$ and $q$ is expressed as $K(q \mid p)$, 
the length of the shortest program
$\bar{w}$ that computes $q$, given $p$, ignoring an additive constant
of $O(1)$ depending on $U$ (in practical applications the computation
will be time-bounded~\cite{LiVitanyi:97}). That is, if $p$ is a
solution to problem $P$, and $q$ is a fast (say, linear time) solution
to problem $Q$, and if $K(q \mid p)$ is small, and $\bar{w}$ is both fast
and much shorter than $q$, then {\em asymptotically optimal universal
  search}~\cite{Levin:73,Schmidhuber:04oops} for a solution to $Q$,
given $p$, will generally find $\bar{w}$ first (to compute $q$ and
solve $Q$), and thus solve $Q$ much faster than search for $q$ from
scratch~\cite{Schmidhuber:04oops}.

In the style of the previous report~\cite{learningtothink2015},
we can directly apply this AIT argument to ONE.
For example, suppose that ONE has learned to represent (e.g., through predictive coding~\cite{chunker91and92,SchmidhuberHeil:96})
 videos of people placing toys in boxes,
or to summarize such videos through textual outputs.
Now suppose ONE's next task is to learn to control a robot that places toys in boxes. 
Although the robot's actuators may be quite different from human arms and hands, 
and although videos and video-describing texts are quite different from desirable trajectories of
robot movements, ONE's knowledge about videos is expected to convey algorithmic 
information about solutions to ONE's new control task, perhaps in form of connected
high-level spatio-temporal  feature detectors representing typical movements of hands and elbows independent of  arm size.  
Training ONE to address this information in its own subroutines
and partially reuse them to solve the robot's task may 
be much faster than learning to solve the task from scratch with a fresh network.

\begin{algorithm}[t]
\begin{algorithmic}
\STATE {\bf 1.} 
Initialize global variables ONE, a finite set  $\cal T$ of task descriptions~\cite[Sec.~2]{powerplay2011and13}, and positive real-valued variables $c, \lambda$ used to define training time budgets.
\STATE {\bf 2.} 
Spend $c$ seconds on trying to solve a 1st task in  $\cal T$ through Algorithm~\ref{ONEalg}, 
then $c$ seconds on trying to solve the 2nd, 
and so on (here a teacher may or may not suggest an initial ordering of tasks).
In line with Algorithm~\ref{ONEalg},
whenever a task gets solved within the allocated time, 
spend  $\lambda c$ seconds on compressing its traces into ONE, 
while also retraining ONE on previous traces to reduce forgetting of older skills,
and even on traces of unsuccessful trials to improve ONE's predictions (if any).
\STATE {\bf 3.} 
If no task in  $\cal T$ got solved, set $c:=2c$ and go to 2.  
\STATE {\bf 4.} 
Set $\cal T$ equal to the set of still unsolved tasks. If $\cal T$ is empty, exit. Reset $c$ to its original value of Step 1. Go to 2 {\em (with a ``more sophisticated" ONE that already knows how to solve some tasks).}
\end{algorithmic}
\caption{Simple automatic ordering of ONE's tasks  - see Sec. \ref{simple}.}
\label{simplealg}
\end{algorithm}

\subsection{Gaining Efficiency by Selective Replays}
\label{selective}

Instead of retraining ONE in a sleep phase
(step 4 of algorithm~\ref{ONEalg}) on all input-output traces of all trials ever,
we may also
retrain it on parts thereof, by selecting trials randomly or
otherwise, and replaying~\cite{Lin:91} them to retrain ONE in standard 
fashion~\cite{learningtothink2015}. 
Generally speaking, we cannot expect perfect compression of previously learned skills and knowledge within limited retraining time spent in a particular invocation of Step 4. 
Nevertheless, repeated incarnations of Step 4 will over time improve ONE's performance on all tasks so far. 

\subsection{Heuristics: Gaining Efficiency by Tracking Weight Variance}
\label{variance}

As a heuristic, we may track the variance of each weight's value at the ends of all trials.
Frequently used weights with low variance can be suspected
to be important for many tasks, and may get small or zero learning rates during Step 3 of Algorithm~\ref{ONEalg},
thus making them even more stable, such that the system does not easily forget them
during the learning of new tasks. 
Weights with high variance, however, may get high learning rates in Step 3, and thus participate 
easily in the learning of new skills.  
Similar heuristics go back to the early days of neural network research.
They can protect ONE's earlier acquired skills and knowledge to a certain extent, to facilitate retraining in Step 4.

\subsection{Gaining Efficiency by Tracking Which Weights Are Used for Which Tasks}
\label{tracking}

To avoid forgetting previous skills,
instead of replaying all previous traces of still relevant trials 
(the simplest option to achieve the  {\sc PowerPlay} criterion~\cite{powerplay2011and13}),
one can also implement ONE as a self-modularizing, computation
cost-minimizing, winner-take-all
RNN~\cite{Schmidhuber:89cs,Schmidhuber:12slimnn,Srivastava2013first}.
Then we can keep track of which weights of ONE are used
 for which tasks.
 That is, to test whether ONE has forgotten something 
 in the wake of recent modifications of some of its weights, 
 only input-output traces in the union of affected tasks  have to
be re-tested~\cite[Sec.~3.3.2]{powerplay2011and13}.
First
implementations of this simple principle were described in previous work on
 {\sc PowerPlay}~\cite{powerplay2011and13,Srivastava2013first}.

\subsection{Ordering Tasks Automatically}
\label{order}

So far the present paper has focused on user-given sequences of tasks. But in general, given a set of tasks, no teacher knows the best sequential ordering of tasks,
to make ONE learn to solve all tasks as quickly as possible. 

The {\sc PowerPlay} framework (2011)~\cite{powerplay2011and13} offers a general solution to the automatic task ordering problem. 
Given is a set of tasks, which may actually be the set of {\em all}  tasks with computable task descriptions, or  a more limited set of tasks, some of them possibly given by a user. In unsupervised mode, one 
{\sc PowerPlay} variant systematically searches the space of possible pairs of new tasks and modifications of the current problem solver, 
until it finds a more powerful problem solver that solves all previously learned tasks plus the new one, while the unmodified predecessor does not. 
The greedy search of typical PowerPlay variants uses time-optimal program search to order candidate pairs of tasks and solver modifications by their conditional computational (time and space) complexity, given the stored experience so far. The new task and its corresponding task-solving skill are those first found and validated. This biases the search toward pairs that can be described compactly and validated quickly. The computational costs of validating new tasks need not grow with task repertoire size.

\subsubsection{Simple automatic ordering of ONE's tasks}
\label{simple}

A related, more naive, but  easy-to-implement strategy is given by {\bf Algorithm~\ref{simplealg}, }
which temporally skips tasks that it currently cannot solve within a given time budget, 
trying to solve them again later after it has learned other skills,
eventually doubling the time budget if any unsolved tasks are left.

\section{Conclusion}

Supervised learning in  large LSTMs works so well that it has become highly commercial, e.g.,~\cite{googlevoice2015,wu2016google,amazon2016,facebook2017}.
True AI, however, must continually learn to solve more and more complex control problems in partially observable environments {\em without a teacher}. 
In principle, this could be achieved by black box optimization through neuroevolution or related techniques.
Such approaches, however, are currently feasible only for networks much smaller than large commercial supervised LSTMs. 
Here we combine the best of both worlds, and apply the AIT argument to 
show how a single recurrent neural network  called ONE can incrementally absorb more and more control and prediction skills 
through rather efficient and well-understood  gradient descent-based compression of desirable behaviors, 
including behaviors of control policies learned by past instances of ONE through neuroevolution or similar general but slow techniques.  
Ideally, none of the ``holy data" from all trials is ever discarded; 
all can be used to incrementally make ONE an increasingly general problem solver
able to solve more and more tasks. 

Essentially, during ONE's dreams, 
gradient-based compression of policies and 
data streams simplifies ONE, 
squeezing the essence of ONE's previously learned skills and knowledge into 
the code implemented within the recurrent weight matrix of  ONE itself.
This can  improve ONE's ability to generalize and quickly learn new, 
related tasks when it is awake.

%\end{sloppypar}

%\newpage
\bibliography{bib}
\bibliographystyle{abbrv}
%\bibliography{bib,bib_extra}
%\bibliographystyle{alpha}
%\bibliographystyle{apalike}
%\printauthorindex
\end{document}